\documentclass[letterpaper]{article} 
\usepackage{aaai2026}  
\usepackage{times}  
\usepackage{helvet} 
\usepackage{courier}  
\usepackage[hyphens]{url}  
\usepackage{graphicx} 
\usepackage{natbib}  
\usepackage{caption} 
\usepackage{algorithm}
\usepackage{algorithmic}
\usepackage[table]{xcolor}
\usepackage{amsmath}
\usepackage{booktabs}
\usepackage{amsfonts}
\usepackage{siunitx}
\usepackage{makecell}
\usepackage{subcaption}
\usepackage{graphicx,overpic}
\newcommand{\modelname}{PathRAG }
\usepackage{newfloat}
\usepackage{listings}
\DeclareCaptionStyle{ruled}{labelfont=normalfont,labelsep=colon,strut=off} 
\floatstyle{ruled}
\newfloat{listing}{tb}{lst}{}
\floatname{listing}{Listing}
\makeatletter
\renewcommand\@fnsymbol[1]{\ensuremath{\ifcase#1\or \dagger\or \ddagger\or
   \mathsection\or \mathparagraph\or \|\or **\or \dagger\dagger
   \or \ddagger\ddagger \else\@ctrerr\fi}}
\makeatother

\title{PathRAG: Pruning Graph-Based Retrieval Augmented Generation with Relational Paths}
\author{
    Boyu Chen\textsuperscript{\rm 1},
    Zirui Guo\textsuperscript{\rm 2},
    Zidan Yang\textsuperscript{\rm 1},
    Yuluo Chen\textsuperscript{\rm 1},
    Junze Chen\textsuperscript{\rm 1},
    Zhenghao Liu\textsuperscript{\rm 3},\\
    Chuan Shi\textsuperscript{\rm 1},
    Cheng Yang\textsuperscript{\rm 1\thanks{Corresponding author.}}
}
\affiliations{
    \textsuperscript{\rm 1}Beijing University of Posts and Telecommunications, China\\
    \textsuperscript{\rm 2}University of Hong Kong, China\\
    \textsuperscript{\rm 3}Northeastern University, China\\
    chenbys4@bupt.edu.cn, yangcheng@bupt.edu.cn
}

\usepackage{bibentry}

\begin{document}

\maketitle

\begin{abstract}
Retrieval-augmented generation (RAG) improves the response quality of large language models (LLMs) by retrieving knowledge from external databases. Typical RAG approaches split the text database into chunks, organizing them in a flat structure for efficient searches. To better capture the inherent dependencies and structured relationships across the text database, researchers propose to organize textual information into an indexing graph, known as \textit{graph-based RAG}. However, we argue that the limitation of current graph-based RAG methods lies in the redundancy of the retrieved information, rather than its insufficiency. Moreover, previous methods use a flat structure to organize retrieved information within the prompts, leading to suboptimal performance. To overcome these limitations, we propose PathRAG, which retrieves key relational paths from the indexing graph, and converts these paths into textual form for prompting LLMs. Specifically, PathRAG effectively reduces redundant information with flow-based pruning, while guiding LLMs to generate more logical and coherent responses with path-based prompting. Experimental results show that PathRAG consistently outperforms state-of-the-art baselines across six datasets and five evaluation dimensions.
\end{abstract}

\begin{links}
    \link{Code}{https://github.com/BUPT-GAMMA/PathRAG}
\end{links}

\section{Introduction}

\begin{figure}[t]
\centering
\includegraphics[width=0.9\columnwidth]{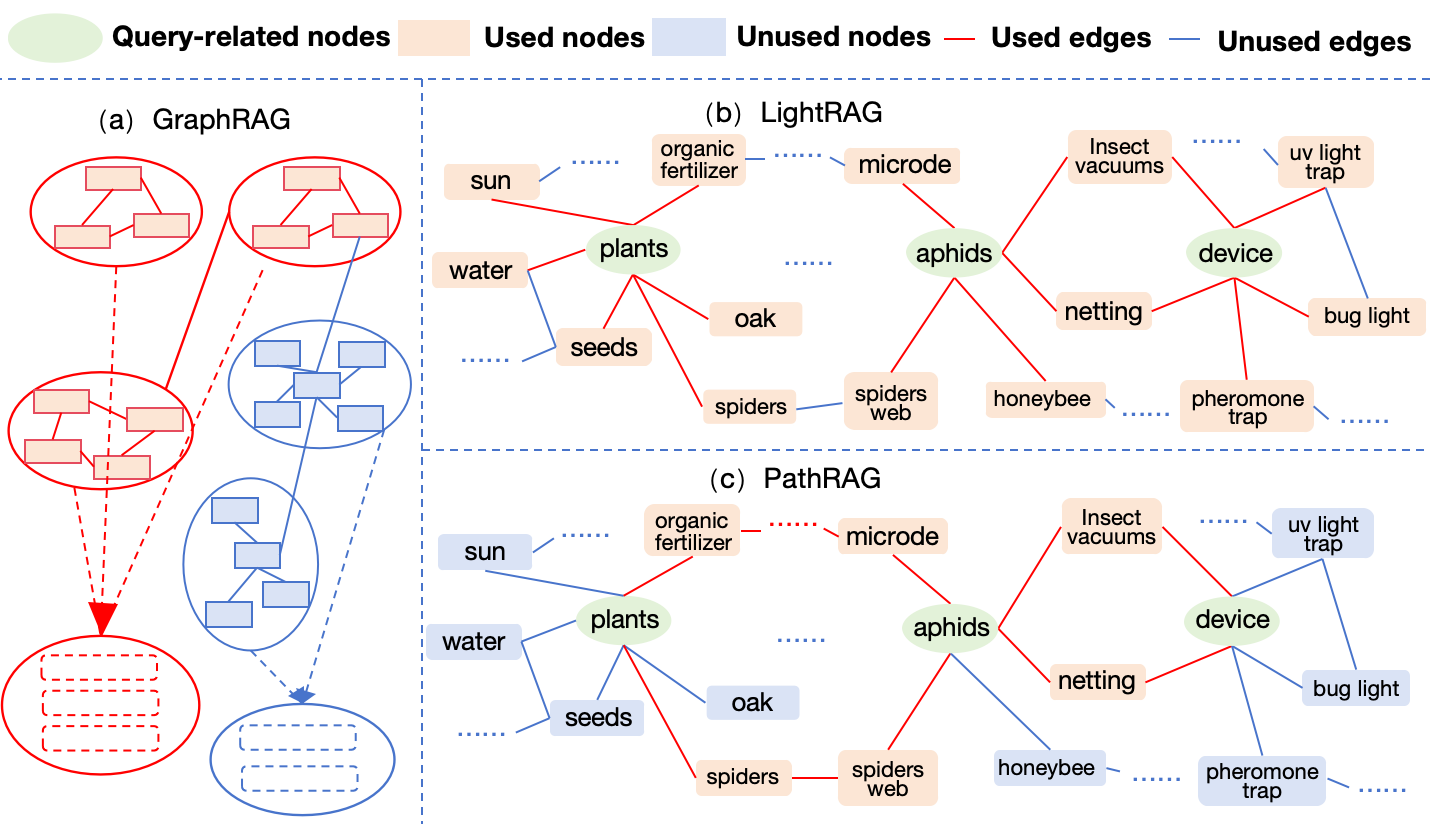} 
\caption{Comparison between different graph-based RAG methods. GraphRAG~\cite{edge2024graphrag} uses all the information within certain communities, while LightRAG~\cite{guo2024lightrag} uses all the immediate neighbors of query-related nodes. In contrast, the proposed PathRAG focuses on key relational paths between query-related nodes to alleviate noise and reduce token consumption. }
\label{fig:Ego graph information vs Path information}
\end{figure}

Retrieval-augmented generation (RAG) empowers large language models (LLMs) to access up-to-date or domain-specific knowledge from external databases, enhancing the response quality without additional training~\cite{gao2022HyDE,gao2023retrieval-naiverag,RAG—survey-2024,kg-survey-2024,mavromatis2024gnn,su2025clue,zhang2025graph}. Most approaches divide the text database into chunks, organizing them in a flat structure to facilitate efficient searches~\cite{yepes2024financial,lyu2024crud}.

To better capture the inherent dependencies and structured relationships across texts in a database, researchers have introduced graph-based RAG~\cite{edge2024graphrag,guo2024lightrag}, which organizes textual information into an indexing graph. In this graph, nodes represent entities extracted from the text, while edges denote the relationships between these entities. Traditional RAG~\cite{liu2021kg,yasunaga2021kg-rag,gao2021kg-rag} usually focuses on questions that can be answered with local information about a single entity or relationship. In contrast, graph-based RAG targets on  global-level questions that need the information across a database to generate a summary-like response. For example, GraphRAG \cite{edge2024graphrag} first applies community detection on the graph, and then gradually summarizes the information in each community. The final answer is generated based on the most query-relevant communities. LightRAG \cite{guo2024lightrag} extracts both local and global keywords from input queries, and retrieves relevant nodes and edges using these keywords. The ego-network information of the retrieved nodes is then used as retrieval results. 

However, we argue that the information considered in previous graph-based RAG methods is often redundant, which can introduce noise, degrade performance, and increase token consumption. 
GraphRAG method uses all the information from the nodes and edges within certain communities. Similarly, LightRAG retrieves the immediate neighbors of query-related nodes to generate answers. The redundant information retrieved in these methods may act as noise, and negatively affecting subsequent generation. Moreover, both methods adopt a flat structure to organize retrieved information in the prompts, \textit{e.g.,} directly concatenating the textual information of all retrieved nodes and edges, resulting in answers with suboptimal logicality and coherence.

To overcome the above limitations, we propose PathRAG, which performs key path retrieval among retrieved nodes and converts these paths into textual form for LLM prompting. 
We focus on the key relational paths between retrieved nodes to alleviate noise and reduce token consumption. Specifically, we first retrieve relevant nodes from the indexing graph based on the keywords in the query. Then we design a flow-based pruning algorithm with distance awareness to identify the key relational paths between each pair of retrieved nodes. The pruning algorithm enjoys low time complexity, and can assign a reliability score to each retrieved path. Afterward, we sequentially concatenate the node and edge information alongside each path as textual relational paths. Considering the ``lost in the middle'' issue of LLMs~\cite{liu2024lost}, we place the textual paths into the prompt in ascending order of reliability scores for better answer generation. 
To comprehensively evaluate the effectiveness of PathRAG, we extend the benchmark datasets from prior work~\cite{guo2024lightrag} with four additional ones~\cite{wang2022squality, chen2022summscreen,qian2024memorag} from different domains. Experimental results demonstrate that PathRAG consistently outperforms state-of-the-art baselines across all five evaluation dimensions. In particular, compared to GraphRAG and LightRAG, PathRAG achieves average win rates of 59.93\% and 57.09\%, respectively.
The contributions of this work are as follows:

$\bullet$ We highlight that the limitation of current graph-based RAG methods lies in the redundancy of the retrieved information, rather than its insufficiency. Moreover, previous methods use a flat structure to organize retrieved information within the prompts, leading to suboptimal performance.

$\bullet$ We propose PathRAG, which efficiently retrieves key relational paths from an indexing graph with flow-based pruning, and effectively generates answers with path-based LLM prompting. 

$\bullet$ PathRAG outperforms state-of-the-art baselines across six datasets and five evaluation dimensions. Extensive experiments further validate the design of PathRAG.

\section{Related Work}
\textbf{Text-based RAG}.
To improve text quality~\cite{fang2024enhancing,xu2024unsupervised,zhu2024information} and mitigate hallucination effects \cite{lewis2020retrieval,guu2020retrieval}, retrieval-augmented generation (RAG) is widely used in large language models (LLMs) by leveraging external databases. These databases primarily store data in textual form, containing a vast amount of domain knowledge that LLMs can directly retrieve. We refer to such systems as text-based RAG. Based on different retrieval mechanisms \cite{RAG—survey-2024}, text-based RAG can be broadly classified into two categories: \textbf{sparse vector retrieval} \cite{alon2022neuro,schick2023toolformer,jiang2023active,cheng2024lift} and \textbf{dense vector retrieval} \cite{lewis2020retrieval,hofstatter2023fid,li2024llama2vec,zhang2024arl2}. Sparse vector retrieval typically identifies the most representative words in each text segment by word frequency, and retrieves relevant text for a specific query based on keyword matching. In contrast, dense vector retrieval addresses issues like lexical mismatches and synonyms by encoding both query terms and text into vector embeddings. It then retrieves relevant content based on the similarity between these embeddings. However, most text-based RAG methods use a flat organization of text segments, and fail to capture essential relationships between chunks (\textit{e.g.,} the contextual dependencies), limiting the quality of LLM-generated responses~\cite{edge2024graphrag,guo2024lightrag}.

\textbf{KG-RAG}. Besides text databases, researchers have proposed retrieving information from knowledge graphs (KGs), known as KG-RAG ~\cite{yasunaga2021kg-rag,gao2021kg-rag,li2024subgraphrag,kg-survey-2024,he2025g}. These methods can utilize existing KGs~\cite{wen2023mindmap,dehghan2024ewek} or their optimized versions~\cite{fang2024reano,panda2024holmes}, and enable LLMs to retrieve information of relevant entities and their relationships. Specifically, KG-RAG methods typically extract a local subgraph from the KG~\cite{bordes2015large,talmor2018web,gu2021beyond}, such as the immediate neighbors of the entity mentioned in a query. However, most KG-RAG methods focus on addressing questions that can be answered with a single entity or relation in the KG~\cite{joshi2017triviaqa,yang2018hotpotqa,kwiatkowski2019natural,ho2020constructing}, narrowing the scope of their applicability.

\textbf{Graph-based RAG}. Instead of utilizing pre-constructed KGs,
graph-based RAG~\cite{edge2024graphrag,guo2024lightrag} typically organizes text databases as text-associated graphs, and focuses on global-level tasks that need the information from multiple segments across a database. The graph construction process often involves extracting entities from the text and identifying relationships between these entities. Also, contextual information is included as descriptive text to minimize the information loss during the text-to-graph conversion. GraphRAG \cite{edge2024graphrag} first applies community detection algorithms on the graph, and then gradually aggregates the information from sub-communities to form higher-level community information. LightRAG \cite{guo2024lightrag} adopts a dual-stage retrieval framework to accelerate the retrieval process. First, it extracts both local and global keywords from the question. Then, it retrieves relevant nodes and edges using these keywords, treating the ego-network information of the retrieved nodes as the final retrieval results. This approach simplifies retrieval and effectively handles global-level tasks. However, the retrieved information covers all immediate neighbors of relevant nodes, which may introduce noise harming the answer quality. A recent work MiniRAG~\cite{fan2025minirag} also considers path information to assist retrieval. But they focus on addressing questions that can be answered by the information of a specific node, and thus explore paths between query-related and answer-related nodes like KG reasoning~\cite{yasunaga2021kg-rag,liu2021kg,tian2022knowledge}.

\section{Preliminaries}
In this section we will introduce and formalize the workflow of a graph-based RAG system.

Instead of storing text chunks as an unordered collection, graph-based RAG automatically structures a text database into an \textbf{indexing graph} as a preprocessing step. Given a text database, the entities and their interrelations within the textual content are identified by LLMs, and utilized to construct the node set \(\mathcal{V}\) and edge set \(\mathcal{E}\). Specifically, each node \(v \in \mathcal{V}\) represents a distinct entity with an identifier \(k_v\) (\textit{e.g.,} entity name) and a textual chunk \(t_v\) (\textit{e.g.,} associated text snippets), while each edge \(e \in \mathcal{E}\) represents the relationship between entity pairs with a descriptive textual chunk \(t_e\) to enrich relational context. We denote the indexing graph as \(\mathcal{\mathcal G = (\mathcal V, \mathcal E, \mathcal K_{\mathcal V}, \mathcal T)}\), where \(\mathcal K_{\mathcal V}\) represent the collection of node identifiers and \(\mathcal T\) is the collection of textual chunks in the indexing graph.

Given a query \( q\), a graph-oriented retriever extracts relevant nodes and edges in the indexing graph. Then the textual chunks of retrieved elements are integrated with query \( q \) to obtain the answer by an LLM generator. The above process can be simplified as:
\begin{equation}
    \mathcal{A}(q, \mathcal G) = \mathcal F \circ \mathcal M (q;\mathcal R(q, \mathcal G)),
\end{equation}
where \( \mathcal A \) denotes the augmented generation with retrieval results, \(\mathcal R\) means the graph-oriented retriever, \( \mathcal M \) and \( \mathcal F \) represent the prompt template and the LLM generator, respectively. In this paper, we primarily focus on designing a more effective graph-oriented retriever and the supporting prompt template to achieve a better graph-based RAG.

\begin{figure*}[t]
    \centering
    \includegraphics[width=0.85\textwidth,height=0.28\textheight]{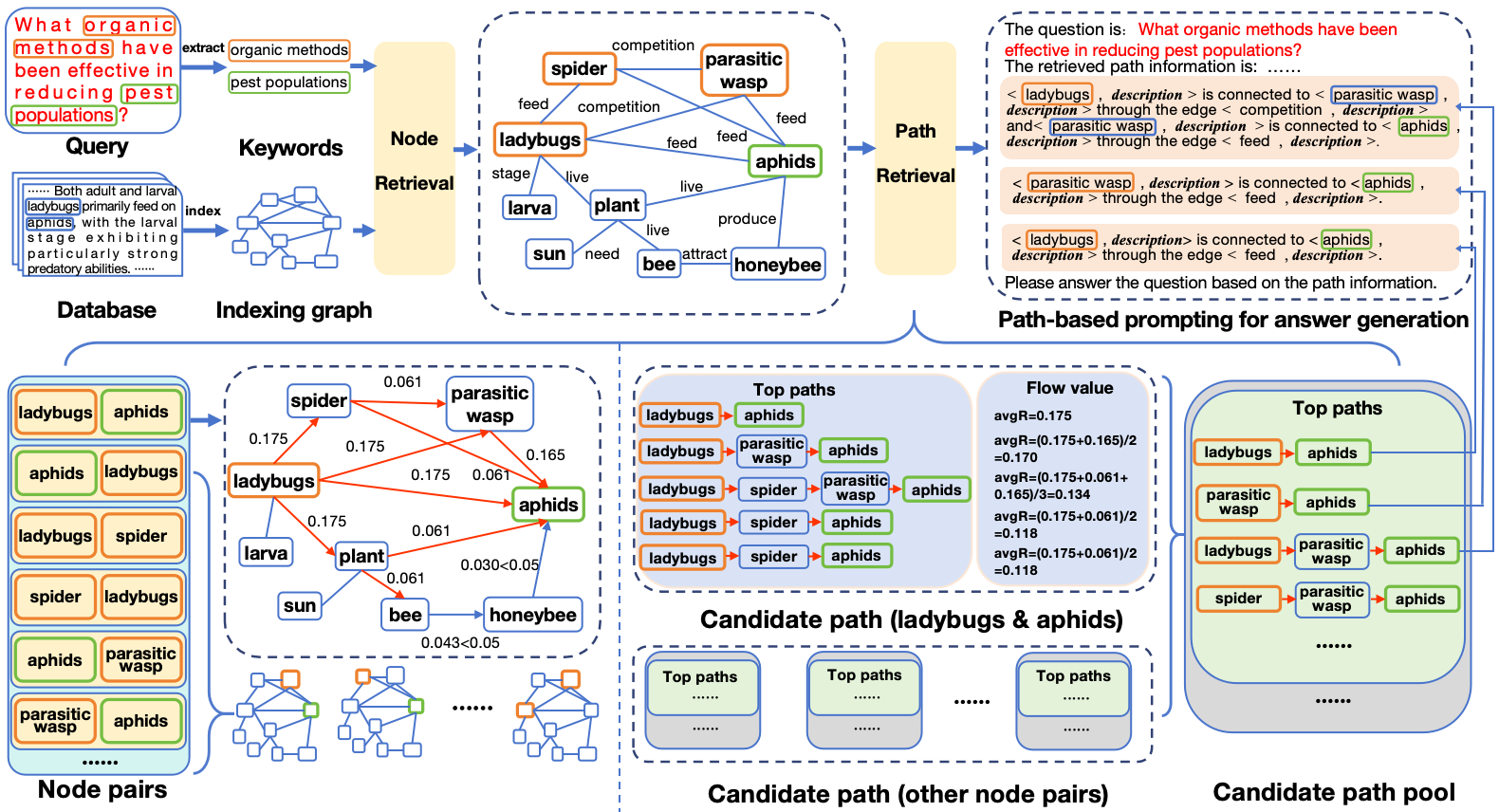}
    \caption{The overall framework of our proposed PathRAG with three main stages. 1) Node Retrieval Stage: Relevant nodes are retrieved from the indexing graph based on the keywords in the query; 2) Path Retrieval Stage: We design a flow-based pruning algorithm to extract key relational paths between each pair of retrieved nodes, and then retrieve paths with the highest reliability scores; 3) Answer Generation Stage: The retrieved paths are placed into prompts in ascending order of reliability scores, and finally fed into an LLM for answer generation.}
    \label{fig:model_overview}
\end{figure*}

\section{Methodology}

In this section, we propose a novel graph-based RAG framework with the path-based retriever and a tailored prompt template, formally designated as PathRAG. As illustrated in Figure \ref{fig:model_overview}, the proposed framework operates on an indexing graph through three sequential stages: node retrieval, path retrieval, and answer generation.

\subsection{Node Retrieval}

In this stage, we identify keywords from the input query by LLMs, and accordingly extract relevant nodes from the indexing graph. Given a query \(q\), an LLM is utilized to extract keywords from the query text. The collection of keywords extracted from query \( q \) is denoted as \( \mathcal{K}_{q} \). Based on the extracted keywords, dense vector matching is employed to retrieve related nodes in the indexing graph \( \mathcal{G} \). In dense vector matching, the relevance between a keyword and a node is calculated by their similarity in the semantic embedding space, where the commonly used cosine similarity is adopted in our method. Specifically, we first encode both node identifiers and the extracted keywords using a semantic embedding model \( f: \mathcal{K}_{q} \cup \mathcal{K}_{\mathcal{V}} \rightarrow \mathcal{X}_{q} \cup \mathcal{X}_{\mathcal{V}} \), where \( \mathcal{X}_{\mathcal{V}} = \{x_{v}\}_{v \in \mathcal{V}} \) represents the embeddings of node identifiers, and \( \mathcal{X}_{q} = \{x_{q,i}\}_{i=1}^{|\mathcal{K}_{q}|} \) denotes the embeddings of the extracted keywords. Based on the obtained embeddings above, we then iterate over \( \mathcal{X}_{q} \) to search the most relevant nodes among \( \mathcal{X}_{\mathcal{V}} \) with the embedding similarity, until a predefined number \( N \) of nodes is reached. The resulting subset of retrieved nodes is denoted as \( \mathcal{V}_{q} \subseteq \mathcal{V} \).

\subsection{Path Retrieval}
In this subsection, we introduce the path retrieval module that aggregates textual chunks in the form of relational paths to capture the connections between retrieved nodes.

Given two distinct retrieved nodes \(v_\text{start}, v_\text{end}\in \mathcal{V}_q\), there could be many reachable paths between them. Since not all paths are helpful to the task, further refinement is needed to enhance both effectiveness and efficiency. 
Inspired by the resource allocation strategy~\cite{lu2011link,lin2015flowing}, we propose a flow-based pruning algorithm with distance awareness to extract key paths.

Formally, we denote the sets of nodes pointing to \(v_i\) and nodes pointed by \(v_i\) as \(\mathcal N(\cdot, v_i)\) and \(\mathcal N(v_i,\cdot)\), respectively. We define the resource of node \(v_i\) as \(\mathcal{S}(v_i)\). We set \(\mathcal{S}(v_\text{start}) = 1\) and initialize other resources to $0$, followed by propagating the resources through the neighborhood. The resource flowing to \(v_i\) is defined as:
\begin{equation}
  \label{eq:flow-algorithm }
  \mathcal{S}(v_i) = \sum_{v_j \in \mathcal N(\cdot,v_i)}\frac{\alpha \cdot \mathcal{S}(v_j)}{\lvert \mathcal N(v_j,\cdot)\rvert},
\end{equation}
where $\alpha$ represents the decay rate of information propagation along the edges. Based on the assumption that the closer two nodes are in the indexing graph, the stronger their connection will be, we introduce this penalty mechanism to enable the retriever to perceive distance. It is crucial to emphasize that our approach differs from strictly sorting paths with a limited number of hops. Detailed comparative experiments will be presented in subsequent sections. 

Notably, due to the decay penalty and neighbor allocation, nodes located far from the initial node are assigned with negligible resources.
Therefore, we introduce an early stopping strategy to prune paths in advance when 
\begin{equation}
    \frac{\mathcal{S}(v_i)}{\lvert\mathcal N(v_i,\cdot) \rvert}<\theta,
\end{equation}
where \(\theta\) is the pruning threshold. This ensures that the algorithm terminates early for nodes that contribute minimally to the overall propagation. For efficiency concerns, we update the resource of a node at most once.

We denote each path as an ordered sequence 
\( P = v_{0} \xrightarrow{e_{0}} \cdots v_{i} \xrightarrow{e_{i}} \cdots  = (\mathcal V_{P}, \mathcal E_{P})\), where \(v_{i}\) and \(e_{i}\) represent the \(i\)-th node and directed edge, and \(\mathcal V_{P}\) and \(\mathcal E_{P}\) represent the set of nodes and edges in the path \(\mathcal P\), respectively. 
For each path $P=(\mathcal{V}_P,\mathcal E_{P})$, we calculate the average resource values flowing through its edges as the measurement of reliability, which can be formulated as:
\begin{equation}
  \label{eq:avg-equation }
  \mathcal S(P) = \frac{1}{{\lvert \mathcal E_{P}\rvert}}\sum_{v_i \in \mathcal{V}_P}S(v_i),
\end{equation}
where \(|\mathcal E_{P}|\) is the number of edges in the path. Then, we sort these paths based on the reliability \(\mathcal S(P)\) and retain only the most reliable relational paths for this node pair. These paths are added to the global candidate pool in the form of path-reliability pair \((P, \mathcal S(P))\). We repeat the above process for each distinct node pair, ultimately obtaining all candidate paths. Then the top-\(K\) reliable paths can be obtained from the candidate pool to serve as the retrieval information of query \(q\) for subsequent generation, which we denote as \(\mathcal P_{q}\).

\subsection{Answer Generation}
For better answer generation, we establish path prioritization based on their reliability, then strategically position these paths to align with LLMs' performance patterns~\cite{qin2023large,liu2024lost,cuconasu2024power}. 

Formally, for each retrieved relational path, we concatenate the textual chunks of all nodes and edges within the path to obtain a textual relational path, which can be formulated as:
\begin{equation}
    t_{P} = \operatorname{concat}([\cdots;t_{v_i}; t_{e_i};t_{v_{i+1}};\cdots]),
\end{equation}
where \(\operatorname{concat}(\cdot)\) denotes the concatenation operation, \(v_i\) and \(e_i\) are the \(i\)-th node and edge in the path \(P\), respectively. 

Considering the ``lost in the middle'' issue~\cite{liu2024lost,cao2024graphinsight} for LLMs in long-context scenarios, directly aggregating the query with different relational paths may lead to suboptimal results. Therefore, we position the most critical information at the two ends of the template, which is regarded as the golden memory region for LLM comprehension. Specifically, we place the query at the beginning of the template and organize the textual relational paths in a reliability ascending order, ensuring that the most reliable relational path is positioned at the end of the template. The final prompt can be denoted as:
\begin{equation}
    \mathcal M (q;\mathcal R(q, \mathcal G))= \operatorname{concat}([q;t_{P_K};\cdots;t_{P_1}]),
\end{equation}
where \(P_1\) is the most reliable path and \(P_K\) is the \(K\)-th reliable path. This simple prompting strategy can significantly improve the response performance of LLM compared with placing the paths in a random or reliability ascending order in our experiments.

\subsection{Discussion}

\textbf{Path Selection Algorithms in RAG.} While previous methods in KG-RAG enhance model reasoning performance through path selection algorithms~\cite{asai2019learning,sun2023think,chen2024plan}, they are primarily designed for tasks that can be answered by the information contained in a single node or edge, making them unsuitable for global-level tasks. Moreover, in prior approaches, paths are typically used as a means to obtain retrieval results rather than being part of the retrieval output itself. When multiple paths lead to retrieval results, no efficient method exists to filter the paths. In contrast, PathRAG employs a flow-based pruning algorithm to evaluate multiple paths between nodes, selecting relevant paths as part of the retrieval result to address global-level tasks.

\textbf{Complexity Analysis of Path Retrieval.} After the $i$-th step of resource propagation, there are at most $\frac{\alpha^i}{\theta}$ nodes alive due to the decay penalty and early stopping. Hence the total number of nodes involved in this propagation is at most $\sum_{i=0}^\infty \alpha^i/\theta=\frac{1}{(1-\alpha)\theta}$. Thus the complexity of extracting candidate paths between all node pairs is $\mathcal{O}(\frac{N^2}{(1-\alpha)\theta})$. In our settings, the number of retrieved nodes $N\in [10,60]$ is much less than the total number of nodes in the indexing graph $|\mathcal{V}|\sim 10^4$. Thus the time complexity is completely acceptable. Further details are provided in Appendix H.

\textbf{Necessity of Path-based Prompting.} Note that different retrieved paths may have shared nodes or edges. To reduce the prompt length, it is possible to flatten the paths and remove duplications as a set of nodes and edges. However, this conversion will lose the semantic relations between the two endpoints of each path. We also validate the necessity of path-based prompting in the experiments.
{
\renewcommand{\arraystretch}{1}
\begin{table*}[ht]
\caption{Performance across six datasets and five evaluation dimensions in terms of win rates.}
\centering
\scriptsize
    \resizebox{\textwidth}{!}{
    \begin{tabular}{lcccccccccccc}
    \hline
     & \multicolumn{2}{c}{\textbf{Legal}} &  \multicolumn{2}{c}{\textbf{History}} & \multicolumn{2}{c}{\textbf{Biology}} & \multicolumn{2}{c}{\textbf{Mix}} & \multicolumn{2}{c}{\textbf{SQuALITY}} & \multicolumn{2}{c}{\textbf{SummScreen}}	  \\ 
     \cline{2-3} \cline{4-5} \cline{6-7} \cline{8-9} \cline{10-11} \cline{12-13}
     & NaiveRAG & \textbf{PathRAG} & NaiveRAG & \textbf{PathRAG} & NaiveRAG & \textbf{PathRAG} & NaiveRAG & \textbf{PathRAG} & NaiveRAG & \textbf{PathRAG} & NaiveRAG & \textbf{PathRAG} 
     \\ 
    \hline
    Comprehensiveness & \cellcolor{gray!30} 31.6\% & \underline{68.4\%} & \cellcolor{gray!30} 33.2\% & \underline{66.8\%} & \cellcolor{gray!30} 29.8\% & \underline{70.2\%} & \cellcolor{gray!30}26.2\% & \underline{73.8\%} & \cellcolor{gray!30}35.2\% & \underline{64.8\%} & \cellcolor{gray!30}30.0\% & \underline{70.0\%} 
    \\
    Diversity & \cellcolor{gray!30}24.4\% & \underline{75.6\%} & \cellcolor{gray!30}38.4\% & \underline{61.6\%} & \cellcolor{gray!30}35.2\% & \underline{64.8\%} & \cellcolor{gray!30}33.2\% & \underline{66.8\%} & \cellcolor{gray!30}29.2\% & \underline{70.8\%} & \cellcolor{gray!30}24.2\% & \underline{75.8\%}
    \\
    Logicality & \cellcolor{gray!30}35.2\% & \underline{64.8\%} & \cellcolor{gray!30}40.4\% & \underline{59.6\%} & \cellcolor{gray!30}34.4\% & \underline{65.6\%} & \cellcolor{gray!30}36.2\% & \underline{63.8\%} & \cellcolor{gray!30}34.4\% & \underline{65.6\%} & \cellcolor{gray!30}30.2\% & \underline{69.8\%}
    \\
    Relevance & \cellcolor{gray!30}27.2\% & \underline{72.8\%} & \cellcolor{gray!30}37.2\% & \underline{62.8\%} & \cellcolor{gray!30}42.0\% & \underline{58.0\%} & \cellcolor{gray!30}38.4\% & \underline{61.6\%} & \cellcolor{gray!30}31.4\% & \underline{68.6\%} & \cellcolor{gray!30}33.0\% & \underline{67.0\%}
    \\
    Coherence & \cellcolor{gray!30}34.0\% & \underline{66.0\%} & \cellcolor{gray!30}42.4\% & \underline{57.6\%} & \cellcolor{gray!30}38.4\% & \underline{61.6\%} & \cellcolor{gray!30}42.0\% & \underline{58.0\%} & \cellcolor{gray!30}37.2\% & \underline{62.8\%} & \cellcolor{gray!30}30.2\% & \underline{69.8\%}
    \\
    \cmidrule{2-3} \cmidrule{4-5} \cmidrule{6-7} \cmidrule{8-9} \cmidrule{10-11} \cmidrule{12-13}
     & HyDE & \textbf{PathRAG} & HyDE & \textbf{PathRAG} & HyDE & \textbf{PathRAG} & HyDE & \textbf{PathRAG} & HyDE & \textbf{PathRAG} & HyDE & \textbf{PathRAG} \\
    \hline
    Comprehensiveness & \cellcolor{gray!30}38.4\% & \underline{61.6\%} & \cellcolor{gray!30}34.8\% & \underline{65.2\%} & \cellcolor{gray!30}33.2\% & \underline{66.8\%} & \cellcolor{gray!30}42.8\% & \underline{57.2\%} & \cellcolor{gray!30}37.2\% & \underline{62.8\%} & \cellcolor{gray!30}30.2\% & \underline{69.8\%}
    \\
    Diversity & \cellcolor{gray!30}21.6\% & \underline{78.4\%} & \cellcolor{gray!30}35.2\% & \underline{64.8\%} & \cellcolor{gray!30}36.0\% & \underline{64.0\%} & \cellcolor{gray!30}33.8\% & \underline{66.2\%} & \cellcolor{gray!30}33.2\% & \underline{66.8\%} & \cellcolor{gray!30}30.0\% & \underline{70.0\%}
    \\
    Logicality & \cellcolor{gray!30}30.2\% & \underline{69.8\%} & \cellcolor{gray!30}38.4\% & \underline{61.6\%} & \cellcolor{gray!30}45.2\% & \underline{54.8\%} & \cellcolor{gray!30}45.6\% & \underline{54.4\%} & \cellcolor{gray!30}35.6\% & \underline{64.4\%} & \cellcolor{gray!30}35.2\% & \underline{64.8\%}
    \\
    Relevance & \cellcolor{gray!30}35.6\% & \underline{64.4\%} & \cellcolor{gray!30}35.6\% & \underline{64.4\%} & \cellcolor{gray!30}46.4\% & \underline{53.6\%} & \cellcolor{gray!30}43.4\% & \underline{56.6\%} & \cellcolor{gray!30}40.4\% & \underline{59.6\%} & \cellcolor{gray!30}31.4\% & \underline{68.6\%}
    \\
    Coherence & \cellcolor{gray!30}42.0\% & \underline{58.0\%} & \cellcolor{gray!30}40.4\% & \underline{59.6\%} & \cellcolor{gray!30}42.4\% & \underline{57.6\%} & \cellcolor{gray!30}45.6\% & \underline{54.4\%} & \cellcolor{gray!30}40.0\% & \underline{60.0\%} & \cellcolor{gray!30}35.6\% & \underline{64.4\%}
    \\
        \cmidrule{2-3} \cmidrule{4-5} \cmidrule{6-7} \cmidrule{8-9} \cmidrule{10-11} \cmidrule{12-13}
     & G-retriever & \textbf{PathRAG} & G-retriever & \textbf{PathRAG} & G-retriever & \textbf{PathRAG} & G-retriever & \textbf{PathRAG} & G-retriever & \textbf{PathRAG} & G-retriever & \textbf{PathRAG} \\
    \hline
    Comprehensiveness & \cellcolor{gray!30}33.8\% & \underline{66.2\%} & \cellcolor{gray!30}41.2\% & \underline{58.8\%} & \cellcolor{gray!30}43.6\% & \underline{56.4\%} & \cellcolor{gray!30}27.4\% & \underline{72.6\%} & \cellcolor{gray!30}35.2\% & \underline{64.8\%} & \cellcolor{gray!30}44.2\% & \underline{55.8\%}
    \\
    Diversity & \cellcolor{gray!30}35.2\% & \underline{64.8\%} & \cellcolor{gray!30}43.6\% & \underline{56.4\%} & \cellcolor{gray!30}32.0\% & \underline{68.0\%} & \cellcolor{gray!30}24.4\% & \underline{75.6\%} & \cellcolor{gray!30}38.4\% & \underline{61.6\%} & \cellcolor{gray!30}30.0\% & \underline{70.0\%} 
    \\
    Logicality & \cellcolor{gray!30}34.4\% & \underline{65.6\%} & \cellcolor{gray!30}42.0\% & \underline{58.0\%} & \cellcolor{gray!30}40.0\% & \underline{60.0\%} & \cellcolor{gray!30}30.2\% & \underline{69.8\%} & \cellcolor{gray!30}40.2\% & \underline{59.8\%} & \cellcolor{gray!30}44.8\% & \underline{55.2\%}
    \\
    Relevance & \cellcolor{gray!30}35.6\% & \underline{64.4\%} & \cellcolor{gray!30}44.0\% & \underline{56.0\%} & \cellcolor{gray!30}38.4\% & \underline{61.6\%} & \cellcolor{gray!30}36.2\% & \underline{63.8\%} & \cellcolor{gray!30}40.0\% & \underline{60.0\%} & \cellcolor{gray!30}41.2\% & \underline{58.8\%}
    \\
    Coherence & \cellcolor{gray!30}38.0\% & \underline{62.0\%} & \cellcolor{gray!30}46.6\% & \underline{53.4\%} & \cellcolor{gray!30}35.2\% & \underline{64.8\%} & \cellcolor{gray!30}34.4\% & \underline{65.6\%} & \cellcolor{gray!30}37.2\% & \underline{62.8\%} & \cellcolor{gray!30}43.6\% & \underline{56.4\%}
    \\
        \cmidrule{2-3} \cmidrule{4-5} \cmidrule{6-7} \cmidrule{8-9} \cmidrule{10-11} \cmidrule{12-13}
     & HippoRAG & \textbf{PathRAG} & HippoRAG & \textbf{PathRAG} & HippoRAG & \textbf{PathRAG} & HippoRAG & \textbf{PathRAG} & HippoRAG & \textbf{PathRAG} & HippoRAG & \textbf{PathRAG} \\
    \hline
    Comprehensiveness & \cellcolor{gray!30}34.4\% & \underline{65.6\%} & \cellcolor{gray!30}43.6\% & \underline{56.4\%} & \cellcolor{gray!30}46.0\% & \underline{54.0\%} & \cellcolor{gray!30}35.8\% & \underline{64.2\%} & \cellcolor{gray!30}43.0\% & \underline{57.0\%} & \cellcolor{gray!30}30.0\% & \underline{70.0\%}
    \\
    Diversity & \cellcolor{gray!30}38.0\% & \underline{62.0\%} & \cellcolor{gray!30}38.4\% & \underline{61.6\%} & \cellcolor{gray!30}24.4\% & \underline{75.6\%} & \cellcolor{gray!30}37.2\% & \underline{62.8\%} & \cellcolor{gray!30}27.2\% & \underline{72.8\%} & \cellcolor{gray!30}26.4\% & \underline{73.6\%} 
    \\
    Logicality & \cellcolor{gray!30}34.4\% & \underline{65.6\%} & \cellcolor{gray!30}45.6\% & \underline{54.4\%} & \cellcolor{gray!30}41.8\% & \underline{58.2\%} & \cellcolor{gray!30}40.2\% & \underline{59.8\%} & \cellcolor{gray!30}47.2\% & \underline{52.8\%} & \cellcolor{gray!30}33.0\% & \underline{67.0\%}
    \\
    Relevance & \cellcolor{gray!30}40.2\% & \underline{59.8\%} & \cellcolor{gray!30}43.2\% & \underline{56.8\%} & \cellcolor{gray!30}40.0\% & \underline{60.0\%} & \cellcolor{gray!30}44.0\% & \underline{56.0\%} & \cellcolor{gray!30}46.6\% & \underline{53.4\%} & \cellcolor{gray!30}34.4\% & \underline{65.6\%}
    \\
    Coherence & \cellcolor{gray!30}41.2\% & \underline{58.8\%} & \cellcolor{gray!30}44.4\% & \underline{55.6\%} & \cellcolor{gray!30}43.6\% & \underline{56.4\%} & \cellcolor{gray!30}45.4\% & \underline{54.6\%} & \cellcolor{gray!30}42.6\% & \underline{57.4\%} & \cellcolor{gray!30}31.8\% & \underline{68.2\%}
    \\
    \cmidrule{2-3} \cmidrule{4-5} \cmidrule{6-7} \cmidrule{8-9} \cmidrule{10-11} \cmidrule{12-13}
     & GraphRAG & \textbf{PathRAG} & GraphRAG & \textbf{PathRAG} & GraphRAG & \textbf{PathRAG} & GraphRAG & \textbf{PathRAG} & GraphRAG & \textbf{PathRAG} & GraphRAG & \textbf{PathRAG} \\
    \hline
    Comprehensiveness & \cellcolor{gray!30}33.8\% & \underline{66.2\%} & \cellcolor{gray!30}41.0\% & \underline{59.0\%} & \cellcolor{gray!30}39.6\% & \underline{60.4\%} & \cellcolor{gray!30}41.2\% & \underline{58.8\%} & \cellcolor{gray!30}42.0\% & \underline{58.0\%} & \cellcolor{gray!30}37.0\% & \underline{63.0\%}
    \\
    Diversity & \cellcolor{gray!30}29.8\% & \underline{70.2\%} & \cellcolor{gray!30}36.6\% & \underline{63.4\%} & \cellcolor{gray!30}38.2\% & \underline{61.8\%} & \cellcolor{gray!30}36.2\% & \underline{63.8\%} & \cellcolor{gray!30}38.4\% & \underline{61.6\%} & \cellcolor{gray!30}41.2\% & \underline{58.8\%} 
    \\
    Logicality & \cellcolor{gray!30}41.6\% & \underline{58.4\%} & \cellcolor{gray!30}43.6\% & \underline{56.4\%} & \cellcolor{gray!30}34.4\% & \underline{65.6\%} & \cellcolor{gray!30}42.0\% & \underline{58.0\%} & \cellcolor{gray!30}42.0\% & \underline{58.0\%} & \cellcolor{gray!30}40.0\% & \underline{60.0\%}
    \\
    Relevance & \cellcolor{gray!30}40.6\% & \underline{59.4\%} & \cellcolor{gray!30}44.0\% & \underline{56.0\%} & \cellcolor{gray!30}42.4\% & \underline{57.6\%} & \cellcolor{gray!30}40.4\% & \underline{59.6\%} & \cellcolor{gray!30}42.4\% & \underline{57.6\%} & \cellcolor{gray!30}44.4\% & \underline{55.6\%}
    \\
    Coherence & \cellcolor{gray!30}38.2\% & \underline{61.8\%} & \cellcolor{gray!30}40.8\% & \underline{59.2\%} & \cellcolor{gray!30}43.6\% & \underline{56.4\%} & \cellcolor{gray!30}41.6\% & \underline{58.4\%} & \cellcolor{gray!30}41.6\% & \underline{58.4\%} & \cellcolor{gray!30}43.6\% & \underline{56.4\%}
    \\
    \cmidrule{2-3} \cmidrule{4-5} \cmidrule{6-7} \cmidrule{8-9} \cmidrule{10-11} \cmidrule{12-13}
     & LightRAG & \textbf{PathRAG} & LightRAG & \textbf{PathRAG} & LightRAG & \textbf{PathRAG} & LightRAG & \textbf{PathRAG} & LightRAG & \textbf{PathRAG} & LightRAG & \textbf{PathRAG} \\
    \hline
    Comprehensiveness & \cellcolor{gray!30}36.6\% & \underline{63.4\%} & \cellcolor{gray!30}44.0\% & \underline{56.0\%} & \cellcolor{gray!30}42.6\% & \underline{57.4\%} & \cellcolor{gray!30}40.4\% & \underline{59.6\%} & \cellcolor{gray!30}44.0\% & \underline{56.0\%} & \cellcolor{gray!30}46.4\% & \underline{53.6\%}
    \\
    Diversity & \cellcolor{gray!30}38.2\% & \underline{61.8\%} & \cellcolor{gray!30}43.2\% & \underline{56.8\%} & \cellcolor{gray!30}43.6\% & \underline{56.4\%} & \cellcolor{gray!30}42.0\% & \underline{58.0\%} & \cellcolor{gray!30}43.2\% & \underline{56.8\%} & \cellcolor{gray!30}46.4\% & \underline{53.6\%}
    \\
    Logicality & \cellcolor{gray!30}37.2\% & \underline{62.8\%} & \cellcolor{gray!30}41.6\% & \underline{58.4\%} & \cellcolor{gray!30}45.2\% & \underline{54.8\%} & \cellcolor{gray!30}43.6\% & \underline{56.4\%} & \cellcolor{gray!30}44.8\% & \underline{55.2\%} & \cellcolor{gray!30}44.8\% & \underline{55.2\%}
    \\
    Relevance & \cellcolor{gray!30}40.0\% & \underline{60.0\%} & \cellcolor{gray!30}44.0\% & \underline{56.0\%} & \cellcolor{gray!30}44.8\% & \underline{55.2\%} & \cellcolor{gray!30}44.0\% & \underline{56.0\%} & \cellcolor{gray!30}45.6\% & \underline{54.4\%} & \cellcolor{gray!30}44.4\% & \underline{55.6\%}
    \\
    Coherence & \cellcolor{gray!30}38.8\% & \underline{61.2\%} & \cellcolor{gray!30}44.4\% & \underline{55.6\%} & \cellcolor{gray!30}44.4\% & \underline{55.6\%} & \cellcolor{gray!30}38.4\% & \underline{61.6\%} & \cellcolor{gray!30}44.4\% & \underline{55.6\%} & \cellcolor{gray!30}46.4\% & \underline{53.6\%}
    \\
    \hline
    \end{tabular}
}
\label{main_results}
\end{table*}
}

\section{Experiments}
We conduct extensive experiments to answer the following research questions (\textbf{RQs}): 
\textbf{RQ1:} How effective is our proposed \modelname compared to the state-of-the-art baselines? 
\textbf{RQ2:} Has each component of our framework played its role effectively? 
\textbf{RQ3:} How does the model perform with indexing graphs of varying sparsity levels?
\textbf{RQ4:} How does our framework perform under different LLM backbones?
\textbf{RQ5:} How much token cost does \modelname require to achieve the performance of state-of-the art baseline?

\subsection{Experimental Setup}
\textbf{Datasets.} We follow the experimental settings of LightRAG~\cite{guo2024lightrag}, and additionally consider four datasets from UltraDomain~\cite{qian2024memorag}, SQuALITY~\cite{wang2022squality} and SummScreen~\cite{chen2022summscreen} for a thorough evaluation. These datasets vary significantly in scale, with token counts ranging from $180,000$ to $5,000,000$. We tune hyperparameters on Agriculture and CS datasets, and then test on the other six datasets.

\textbf{Baselines.} We compare PathRAG with six state-of-the-art methods: NaiveRAG~\cite{gao2023retrieval-naiverag}, HyDE~\cite{gao2022HyDE}, {G-retriever~\cite{he2025g}, HippoRAG~\cite{gutierrez2024hipporag}}, GraphRAG \cite{edge2024graphrag}, and LightRAG \cite{guo2024lightrag}. These methods cover cutting-edge text-based, {KG-based} and graph-based RAG approaches. 

\textbf{Implementation Details.} 
{To ensure fairness in the experimental process, we uniformly use ``GPT-4o-mini'' as the base model for all methods, and adopt the ``text-embedding-3-small'' model for embedding. In addition, we construct the indexing graph following the method of GraphRAG~\cite{edge2024graphrag}, and the retrieved edges corresponding to the global keywords in LightRAG~\cite{guo2024lightrag} are placed after the query. For components involving randomness, we average over ten trials. The maximum input token length for the LLMs is limited to $8,000$ to ensure fair handling of different forms of retrieved information. The hyperparameters of PathRAG are fixed as $N=40$, $K=15$, and $\alpha=0.7$. }

\renewcommand{\arraystretch}{1}
\begin{table*}[ht]
\caption{Ablation study on the path retrieval algorithm of PathRAG.}
\scriptsize
\centering
    \resizebox{\textwidth}{!}{
    \begin{tabular}
    {lcccccccccccc}
    \hline
     & \multicolumn{2}{c}{\textbf{Legal}} &  \multicolumn{2}{c}{\textbf{History}} & \multicolumn{2}{c}{\textbf{Biology}} & \multicolumn{2}{c}{\textbf{Mix}} & \multicolumn{2}{c}{\textbf{SQuALITY}} & \multicolumn{2}{c}{\textbf{SummScreen}}	  \\ 
     \cline{2-3} \cline{4-5} \cline{6-7} \cline{8-9} \cline{10-11} \cline{12-13}
     & Random & \textbf{Flow-based} & Random & \textbf{Flow-based} & Random & \textbf{Flow-based} & Random & \textbf{Flow-based} & Random & \textbf{Flow-based} & Random & \textbf{Flow-based} \\
    \hline
    Comprehensiveness & \cellcolor{gray!30}44.0\% & \underline{56.0\%} & \cellcolor{gray!30}46.0\% & \underline{54.0\%} & \cellcolor{gray!30}44.0\% & \underline{56.0\%} & \cellcolor{gray!30}42.8\% & \underline{57.2\%} & \cellcolor{gray!30}45.6\% & \underline{54.4\%} & \cellcolor{gray!30}46.6\% & \underline{53.4\%}
    \\ 
    Diversity & \cellcolor{gray!30}45.2\% & \underline{54.8\%} & \cellcolor{gray!30}31.4\% & \underline{68.6\%} & \cellcolor{gray!30}29.8\% & \underline{70.2\%} & \cellcolor{gray!30}46.0\% & \underline{54.0\%} & \cellcolor{gray!30}44.4\% & \underline{55.6\%} & \cellcolor{gray!30}42.8\% & \underline{57.2\%}
    \\ 
    Logicality & \cellcolor{gray!30}46.6\% & \underline{53.4\%} & \cellcolor{gray!30}44.0\% & \underline{56.0\%} & \cellcolor{gray!30}42.0\% & \underline{58.0\%} & \cellcolor{gray!30}46.4\% & \underline{53.6\%} & \cellcolor{gray!30}41.8\% & \underline{58.2\%} & \cellcolor{gray!30}43.6\% & \underline{56.4\%}
    \\ 
    Relevance & \cellcolor{gray!30}44.8\% & \underline{55.2\%} & \cellcolor{gray!30}46.0\% & \underline{54.0\%} & \cellcolor{gray!30}45.8\% & \underline{54.2\%} & \cellcolor{gray!30}45.6\% & \underline{54.4\%} & \cellcolor{gray!30}45.6\% & \underline{54.4\%} & \cellcolor{gray!30}44.8\% & \underline{55.2\%} 
    \\ 
    Coherence & \cellcolor{gray!30}44.6\% & \underline{55.4\%} & \cellcolor{gray!30}41.0\% & \underline{59.0\%} & \cellcolor{gray!30}41.6\% & \underline{58.4\%} & \cellcolor{gray!30}44.0\% & \underline{56.0\%} & \cellcolor{gray!30}43.6\% & \underline{56.4\%} & \cellcolor{gray!30}46.4\% & \underline{53.6\%}
    \\ 
    \cmidrule{2-3} \cmidrule{4-5} \cmidrule{6-7} \cmidrule{8-9} \cmidrule{10-11} \cmidrule{12-13}
     & Hop-first & \textbf{Flow-based} & Hop-first & \textbf{Flow-based} & Hop-first & \textbf{Flow-based} & Hop-first & \textbf{Flow-based} & Hop-first & \textbf{Flow-based} & Hop-first & \textbf{Flow-based} \\
    \hline
    Comprehensiveness & \cellcolor{gray!30}44.4\% & \underline{55.6\%} & \cellcolor{gray!30}45.8\% & \underline{54.2\%} & \cellcolor{gray!30}48.8\% & \underline{51.2\%} & \cellcolor{gray!30}43.2\% & \underline{56.8\%} & \cellcolor{gray!30}45.2\% & \underline{54.8\%} & \cellcolor{gray!30}46.4\% & \underline{53.6\%}
    \\ 
    Diversity & \cellcolor{gray!30}36.0\% & \underline{64.0\%} & \cellcolor{gray!30}49.6\% & \underline{50.4\%} & \cellcolor{gray!30}46.0\% & \underline{54.0\%} & \cellcolor{gray!30}47.6\% & \underline{52.4\%} & \cellcolor{gray!30}44.0\% & \underline{56.0\%} & \cellcolor{gray!30}45.4\% & \underline{54.6\%}
    \\ 
    Logicality & \cellcolor{gray!30}45.2\% & \underline{54.8\%} & \cellcolor{gray!30}41.2\% & \underline{58.8\%} & \cellcolor{gray!30}44.8\% & \underline{55.2\%} & \cellcolor{gray!30}43.6\% & \underline{56.4\%} & \cellcolor{gray!30}46.0\% & \underline{54.0\%} & \cellcolor{gray!30}46.0\% & \underline{54.0\%}
    \\ 
    Relevance & \cellcolor{gray!30}43.6\% & \underline{56.4\%} & \cellcolor{gray!30}46.0\% & \underline{54.0\%} & \cellcolor{gray!30}37.4\% & \underline{62.6\%} & \cellcolor{gray!30}41.4\% & \underline{58.6\%} & \cellcolor{gray!30}44.8\% & \underline{55.2\%} & \cellcolor{gray!30}46.8\% & \underline{53.2\%}
    \\ 
    Coherence & \cellcolor{gray!30}41.0\% & \underline{59.0\%} & \cellcolor{gray!30}40.0\% & \underline{60.0\%} & \cellcolor{gray!30}42.6\% & \underline{57.4\%} & \cellcolor{gray!30}44.8\% & \underline{55.2\%} & \cellcolor{gray!30}46.8\% & \underline{53.2\%} & \cellcolor{gray!30}46.4\% & \underline{53.6\%}
    \\ 
    \hline
    \end{tabular}
}
\label{table:sorting algorithm}
\end{table*}

\renewcommand{\arraystretch}{0.9}
\begin{table*}[ht]
\caption{Ablation study on the prompt format of PathRAG.}
\scriptsize
    \centering
    \resizebox{\textwidth}{!}{
    \begin{tabular}
    {lcccccccccccc}
    \hline
     & \multicolumn{2}{c}{\textbf{Legal}} &  \multicolumn{2}{c}{\textbf{History}} & \multicolumn{2}{c}{\textbf{Biology}} & \multicolumn{2}{c}{\textbf{Mix}} & \multicolumn{2}{c}{\textbf{SQuALITY}} & \multicolumn{2}{c}{\textbf{SummScreen}}	  \\
     \cline{2-3} \cline{4-5} \cline{6-7} \cline{8-9} \cline{10-11} \cline{12-13}
     & Flat & \textbf{Path-based} & Flat & \textbf{Path-based} & Flat & \textbf{Path-based} & Flat & \textbf{Path-based} & Flat & \textbf{Path-based} & Flat & \textbf{Path-based} \\
    \hline
    Comprehensiveness & \cellcolor{gray!30}40.0\% & \underline{60.0\%} & \cellcolor{gray!30}48.8\% & \underline{51.2\%} & \cellcolor{gray!30}45.6\% & \underline{54.4\%} & \cellcolor{gray!30}49.6\% & \underline{50.4\%} & \cellcolor{gray!30}47.2\% & \underline{52.8\%} & \cellcolor{gray!30}46.8\% & \underline{53.2\%}
    \\ 
    Diversity & \cellcolor{gray!30}42.0\% & \underline{58.0\%} & \cellcolor{gray!30}39.6\% & \underline{60.4\%} & \cellcolor{gray!30}44.4\% & \underline{55.6\%} & \cellcolor{gray!30}43.2\% & \underline{56.8\%} & \cellcolor{gray!30}44.4\% & \underline{55.6\%} & \cellcolor{gray!30}45.2\% & \underline{54.8\%}
    \\ 
    Logicality & \cellcolor{gray!30}37.2\% & \underline{62.8\%} & \cellcolor{gray!30}45.6\% & \underline{54.4\%} & \cellcolor{gray!30}48.0\% & \underline{52.0\%} & \cellcolor{gray!30}42.0\% & \underline{58.0\%} & \cellcolor{gray!30}47.6\% & \underline{52.4\%} & \cellcolor{gray!30}43.2\% & \underline{56.8\%}
    \\ 
    Relevance & \cellcolor{gray!30}44.8\% & \underline{55.2\%} & \cellcolor{gray!30}49.0\% & \underline{51.0\%} & \cellcolor{gray!30}47.4\% & \underline{52.6\%} & \cellcolor{gray!30}44.8\% & \underline{55.2\%} & \cellcolor{gray!30}45.4\% & \underline{54.6\%}  & \cellcolor{gray!30}46.0\% & \underline{54.0\%}
    \\ 
    Coherence & \cellcolor{gray!30}39.2\% & \underline{60.8\%} & \cellcolor{gray!30}45.6\% & \underline{54.4\%} & \cellcolor{gray!30}44.6\% & \underline{55.4\%} & \cellcolor{gray!30}42.4\% & \underline{57.6\%} & \cellcolor{gray!30}48.0\% & \underline{52.0\%} & \cellcolor{gray!30}46.8\% & \underline{53.2\%}
    \\ 
    \hline
    \end{tabular}
}
\label{table:prompt structure}
\end{table*}

\textbf{Evaluation Metrics.} Due to the absence of ground truth answers, we follow the LLM-based evaluation procedures as GraphRAG and LightRAG. Specifically, we utilize ``GPT-4o-mini'' to evaluate the generated answers across multiple dimensions. The evaluation dimensions are based on those from GraphRAG and LightRAG, including Comprehensiveness and Diversity, while also incorporating three new dimensions from recent advances in LLM-based evaluation~\cite{chan2023chateval}, namely Logicality, Relevance, and Coherence. We compare the answers generated by each baseline and our method and conduct win-rate statistics. A higher win rate indicates a greater performance advantage over the other. Note that the presentation order of two answers will be alternated, and the average win rates will be reported. More experimental setup details are provided in Appendices A, B, C, and D.

\subsection{Main Results (RQ1)}

As shown in Table~\ref{main_results}, \textbf{PathRAG consistently outperforms the baselines across all evaluation dimensions and datasets}. From the perspective of evaluation dimensions, compared to all baselines, PathRAG shows an average win rate of {62.52\%} in Comprehensiveness, {65.37\%} in Diversity, {60.68\%} in Logicality, {59.92\%} in Relevance, and {59.43\%} in Coherence on average. These advantages highlight the effectiveness of our proposed path-based retrieval, which contributes to better performance across multiple aspects of the generated responses. 
From a dataset-level perspective, PathRAG achieves notable average win rates of 64.66\%, 58.94\%, 60.44\%, and 61.59\% on the Legal, History, Biology, and Mix datasets, respectively. Furthermore, it demonstrates robust performance on the more challenging and unconventional SQuALITY and SummScreen benchmarks, with average win rates of 60.67\% and 63.20\%. These results collectively indicate that PathRAG offers superior multi-domain adaptability and consistently outperforms baseline models across diverse evaluation scenarios.

Considering the human-written summaries in the SQuALITY dataset, we further evaluate the alignment between generated answers and references using automated metrics such as BLEU, ROUGE, and METEOR. As shown in Table~\ref{tab:Automated-evaluation}, PathRAG achieves superior performance across all metrics, with a 7.06\% average improvement over the best baseline. In future work, we will also integrate human evaluation and other semantic-level assessment methods.

\begin{table}[ht!]
    \caption{Evaluation on the SQuALITY dataset using human-written summaries.}
    \centering
    \resizebox{\columnwidth}{!}{
        \begin{tabular}{lccccc}
            \toprule
             & \textbf{BLEU-1} & \textbf{BLEU-2} & \textbf{ROUGE-1-F1} & \textbf{ROUGE-2-F1} & \textbf{METEOR} \\
            \midrule
            NaiveRAG & 31.78\% & 12.31\% & 13.80\% & 3.51\% & 16.90\% \\
            HyDE & 31.68\% & 11.84\% & 13.95\% & 3.50\% & 16.95\% \\
            G-retriever & 32.42\% & 12.03\% & 14.02\% &3.12\% & 17.50\% \\
            HippoRAG & 32.12\% & 11.89\% & 14.18\% & 3.39\% & 17.61\% \\
            GraphRAG & 32.98\% & 12.27\% & 14.23\% & 3.59\% & 17.52\% \\
            LightRAG & 33.37\% & 12.42\% & 14.56\% & 3.30\% & 17.66\% \\
            PathRAG & \underline{35.41\%} & \underline{13.81\%} & \underline{15.35\%} & \underline{3.95\%} & \underline{18.53\%} \\
            \bottomrule
        \end{tabular}
    }
    \label{tab:Automated-evaluation}
\end{table}

\subsection{Ablation Study (RQ2)}

We conduct ablation experiments to validate the design of PathRAG. A detailed introduction to the variants can be found in Appendix G.

\textbf{Necessity of Path Ordering}. We consider two different strategies to rank the retrieved paths in the prompt, namely random and hop-first. As shown in the Table ~\ref{table:sorting algorithm}, the average win rates of PathRAG compared to the random and hop-first variants are respectively 56.44\% and 55.64\%, indicating the necessity of path ordering in the prompts. 

\textbf{Necessity of Path-based Prompting}. 
While retrieval is conducted using paths, the retrieved information in the prompts does not necessarily need to be organized in the same manner. To assess the necessity of path-based organization, we compare prompts structured by paths with those using a flat organization. As shown in Table~\ref{table:prompt structure}, path-based prompts achieve an average win rate of {55.19\%}, outperforming the flat format. In PathRAG, node and edge information within a path is inherently interconnected, and separating them can result in information loss. Therefore, after path retrieval, prompts should remain structured to preserve contextual relationships and enhance answer quality.

\subsection{Graph Sparsity Analysis (RQ3)}
\begin{figure}[ht]
    \centering    
    \includegraphics[width=\columnwidth,height=0.18\textheight]{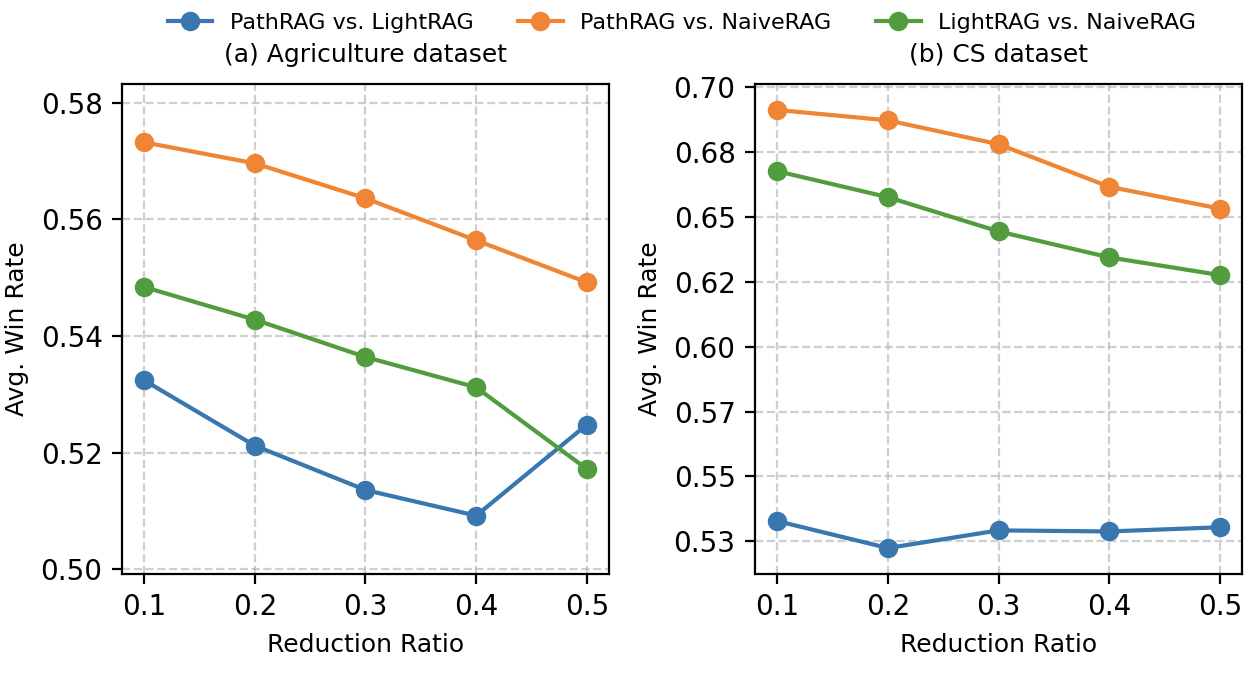}
    \caption{Performance of PathRAG, LightRAG, and NaiveRAG under different levels of graph sparsity on the Agriculture and CS datasets.}
    \label{fig:Sparsity analysis}
\end{figure}

{To assess the robustness of PathRAG under varying levels of graph sparsity, we conduct experiments on the Agriculture and CS datasets. We simulate different sparsity levels by randomly removing 10\%, 20\%, 30\%, 40\%, and 50\% of the edges from the original indexing graphs. Subsequently, we perform pairwise comparisons among PathRAG, LightRAG, and NaiveRAG under each sparsity condition. The results of this evaluation are presented in Figure~\ref{fig:Sparsity analysis}.}

\textbf{Although graph-based methods exhibit a degree of performance degradation when applied to increasingly sparse indexing graphs, PathRAG continues to demonstrate practical robustness.} The simulation of sparsity through the random removal of a substantial proportion of edges does lead to a measurable decline in effectiveness. Nevertheless, PathRAG consistently outperforms both LightRAG, which similarly relies on graph structure and NaiveRAG, which operates independently of any graph-based context. Specifically, on the Agriculture and CS datasets, PathRAG achieves average win rates ranging from 54.84\% to 57.32\% and from 51.72\% to 54.92\%, respectively, when compared with NaiveRAG. Against LightRAG, PathRAG maintains win rates between 50.92\% and 53.24\% on Agriculture, and between 52.24\% and 53.28\% on CS. These findings underscore the robustness and generalizability of PathRAG, even under conditions characterized by severe graph sparsity.

\subsection{Performance under Different LLMs (RQ4)}

\begin{table}[ht!]
\centering
\caption{Performance comparison of PathRAG and LightRAG across different base and evaluation models.}
\renewcommand{\arraystretch}{1} 
    \resizebox{\columnwidth}{!}{
    \begin{tabular}
    {l ccccc}
    \hline
     & \multicolumn{2}{c}{\textbf{Agriculture}} &  \multicolumn{2}{c}{\textbf{CS}} 	  \\
     \cline{2-3} \cline{4-5} 
     \textbf{GPT-4o-mini} & LightRAG & \textbf{PathRAG} & LightRAG & \textbf{PathRAG}  \\
    \hline
    Comprehensiveness & \cellcolor{gray!30}47.6\% & \underline{52.4\%} & \cellcolor{gray!30}47.6\% & \underline{52.4\%} 
    \\ 
    Diversity & \cellcolor{gray!30}44.4\% & \underline{55.6\%} & \cellcolor{gray!30}42.6\% & \underline{57.4\%} 
    \\ 
    Logicality & \cellcolor{gray!30}48.0\% & \underline{52.0\%} & \cellcolor{gray!30}42.6\% & \underline{57.4\%} 
    \\ 
    Relevance & \cellcolor{gray!30}46.6\% & \underline{53.4\%} & \cellcolor{gray!30}45.2\% & \underline{54.8\%} 
    \\ 
    Coherence & \cellcolor{gray!30}45.6\% & \underline{54.4\%} & \cellcolor{gray!30}47.2\% & \underline{52.8\%} 
    \\ 
    \hline
         \textbf{GPT-4o} & LightRAG & \textbf{PathRAG} & LightRAG & \textbf{PathRAG}  \\
    \hline
    Comprehensiveness & \cellcolor{gray!30}47.4\% & \underline{52.6\%} & \cellcolor{gray!30}32.8\% & \underline{67.2\%} 
    \\ 
    Diversity & \cellcolor{gray!30}43.2\% & \underline{56.8\%} & \cellcolor{gray!30}34.0\% & \underline{66.0\%} 
    \\ 
    Logicality & \cellcolor{gray!30}45.6\% & \underline{54.4\%} & \cellcolor{gray!30}42.4\% & \underline{57.6\%} 
    \\ 
    Relevance & \cellcolor{gray!30}42.0\% & \underline{58.0\%} & \cellcolor{gray!30}41.6\% & \underline{58.4\%} 
    \\ 
    Coherence & \cellcolor{gray!30}42.2\% & \underline{57.8\%} & \cellcolor{gray!30}45.2\% & \underline{54.8\%} 
    \\ 
    \hline
         \textbf{DeepSeek-V3} & LightRAG & \textbf{PathRAG} & LightRAG & \textbf{PathRAG}  \\
    \hline
    Comprehensiveness & \cellcolor{gray!30}47.4\% & \underline{52.6\%} & \cellcolor{gray!30}46.0\% & \underline{54.0\%} 
    \\ 
    Diversity & \cellcolor{gray!30}42.0\% & \underline{58.0\%} & \cellcolor{gray!30}42.4\% & \underline{57.6\%} 
    \\ 
    Logicality & \cellcolor{gray!30}44.0\% & \underline{56.0\%} & \cellcolor{gray!30}42.0\% & \underline{58.0\%} 
    \\ 
    Relevance & \cellcolor{gray!30}44.4\% & \underline{55.6\%} & \cellcolor{gray!30}42.2\% & \underline{57.8\%} 
    \\ 
    Coherence & \cellcolor{gray!30}43.2\% & \underline{56.8\%} & \cellcolor{gray!30}41.6\% & \underline{58.4\%} 
    \\ 
    \hline
    \end{tabular}
}
\label{table:model robustness}
\end{table}

Considering that LLMs with different performance levels may affect the overall effectiveness of the framework and the reliability of evaluation, we uniformly replace ``GPT-4o-mini'' with either ``GPT-4o'' or ``DeepSeek-V3'' as the base and evaluation models for comparative experiments. As shown in Table~\ref{table:model robustness}, when ``GPT-4o-mini'' is used as both the base and evaluation model, PathRAG achieves an average win rate of 53.92\% on the Agriculture and CS datasets. When the ``DeepSeek-V3'' model is used, the average win rate increases to 56.48\%. With the highest-performing model, ``GPT-4o'', the win rate reaches a peak of 58.36\%. These results indicate that the stronger the LLM used, the better the overall framework performance, and that PathRAG maintains stable performance across models with varying capabilities.

\subsection{Token Cost Analysis (RQ5)}

For a fair comparison focusing on token consumption, we also consider a lightweight version of PathRAG with $N=20$ and $K=5$, dubbed as PathRAG-lt. PathRAG-lt performs on par with LightRAG in overall performance, achieving an average win rate of 50.56\%, with detailed results provided in the Appendix I.

\begin{table}[ht!]
    \caption{Comparison of LightRAG, PathRAG-lt and PathRAG in terms of token, time, and monetary cost.}
    \centering
    \renewcommand{\arraystretch}{1} 
    \resizebox{\columnwidth}{!}{
        \begin{tabular}{lccc}
            \toprule
             & \textbf{LightRAG} & \textbf{PathRAG-lt} & \textbf{PathRAG} \\
            \midrule
            \textbf{token cost} & 16,728 & 9,968 & 14,438\\
            \textbf{monetary cost} & $2.51 \times 10^{-3}\$ $ & $1.50 \times 10^{-3}\$$ & $2.17 \times 10^{-3}\$$ \\
            \bottomrule
        \end{tabular}
    }
    \label{table:cost of PathRAG}
\end{table}

As shown in Table~\ref{table:cost of PathRAG}, PathRAG achieves significantly better performance while reducing token consumption by 13.69\%, with a corresponding cost of only $0.002\$$. Meanwhile, PathRAG-lt reduces token usage by 40.41\% while maintaining similar performance to LightRAG. These results demonstrate the token efficiency of our method.

\section{Conclusion}
In this paper, we propose PathRAG, a novel graph-based RAG method that focuses on retrieving key relational paths from the indexing graph to alleviate noise. PathRAG can efficiently identify key paths with a flow-based pruning algorithm, and effectively generate answers with path-based LLM prompting. Experimental results demonstrate that PathRAG consistently outperforms baseline methods on six datasets. In future work, we will optimize the indexing graph construction process, and consider to collect more human-annotated datasets for graph-based RAG. It is also possible to explore other substructures besides paths to enhance model performance. 

\section{Acknowledgments}
This work is supported in part by the National Natural Science Foundation of China  (No.62192784, 62236003, 62576082), Young Elite Scientists Sponsorship Program (No.2023QNRC001) by CAST, and Beijing Natural Science Foundation (No.L253004).

\bibliography{aaai2026}
\end{document}